\documentclass{article} 
\usepackage{times}
\usepackage[letterpaper,textwidth=5.5in,textheight=9in]{geometry}
\usepackage{natbib}

\usepackage{amsmath,amsfonts,bm}

\def\eqref#1{equation~\ref{#1}}
\def\Eqref#1{Equation~\ref{#1}}

\def\1{\bm{1}}

\def\rvz{{\mathbf{z}}}

\def\vzero{{\bm{0}}}

\def\vmu{{\bm{\mu}}}

\def\vh{{\bm{h}}}

\def\vu{{\bm{u}}}

\def\vx{{\bm{x}}}

\def\mI{{\bm{I}}}

\DeclareMathAlphabet{\mathsfit}{\encodingdefault}{\sfdefault}{m}{sl}
\SetMathAlphabet{\mathsfit}{bold}{\encodingdefault}{\sfdefault}{bx}{n}

\def\sC{{\mathbb{C}}}
\def\sD{{\mathbb{D}}}

\def\sS{{\mathbb{S}}}
\def\sT{{\mathbb{T}}}

\def\sX{{\mathbb{X}}}
\def\sY{{\mathbb{Y}}}

\newcommand{\R}{\mathbb{R}}

\usepackage{url}
\usepackage{booktabs}
\usepackage{float}
\usepackage{graphicx}
\usepackage{placeins}
\usepackage{capt-of}
\usepackage[
    colorlinks=true,
    linkcolor=blue,
    citecolor=blue,
    urlcolor=blue
]{hyperref}

\title{Amortized Variational Inference of Latent Intensities from Replicated Point Patterns}

\author{\parbox{\textwidth}{\centering
Wenqing Liu\textsuperscript{*1}, Xenia Miscouridou\textsuperscript{\textdagger 2}, and D\'eborah Sulem\textsuperscript{*1}\\[0.8em]
\small
\textsuperscript{1}Faculty of Informatics, Universit\`a della Svizzera italiana, Lugano, Switzerland\\[0.3em]
\textsuperscript{2}Department of Mathematics and Statistics, University of Cyprus, Nicosia, Cyprus
}}

\date{}
\begin{document}

\maketitle

\begin{abstract}
Bayesian inference for latent point process intensities often relies on per-dataset sampling or optimization, making repeated inference computationally expensive. We propose a neural-network-based amortized variational framework that learns a reusable inference rule mapping a variable number of replicated point patterns to a distribution over their shared latent intensity. Trained from event times using the point process likelihood, the model learns an amortized inference 
rule without ground-truth intensity supervision. 
At test time, it performs inference of the latent distribution in a forward pass. On synthetic 
data, additional realizations improve intensity recovery and yield narrower credible intervals with broadly stable coverage. Our method is tested on bicycle-sharing data and achieves the highest mean held-out log likelihood among the evaluated methods while requiring substantially less inference time than the Log-Gaussian Cox Process baseline.

\end{abstract}

\section{Introduction}
\label{sec:intro}
Temporal point processes~\citep{daley2003introduction} provide a general
framework for modeling events that occur in time, with
applications ranging from earthquakes~\citep{ogata1988statistical,ogata1998space}
and neural spikes~\citep{amemori2001gaussian} to criminal activity~\citep{mohler2011self}.
They are commonly characterized by an intensity
function that describes how the event rate varies over time. Cox processes~\citep{moller1998log}
extend this framework by treating the intensity itself as
a latent random function. Bayesian inference quantifies uncertainty over this
function from observed events. Inference in this setting includes
Markov chain Monte Carlo (MCMC)~\citep{adams2009tractable},
integrated nested Laplace approximations (INLA)~\citep{rue2009approximate}, and variational approximations~\citep{lloyd2015variational,donner2018efficient}.
These approaches generally compute a posterior distribution per dataset, which can be costly when
many distinct data sources must be analyzed.

We consider data consisting of repeated temporal point-pattern observations. A temporal
\emph{point pattern} is the sequence of event times
observed within a time window:
$
\sS_{\tau,k}
=
\{t_{\tau,k,1},\ldots,t_{\tau,k,N_{\tau,k}}\}.
$

We group point patterns that are assumed to share the same underlying
intensity into a batch and call this a \emph{task}. Thus, a task $\tau$ consists of
$K_\tau \geq 1$ replicated point patterns,
$
\sD_\tau
=
\{\sS_{\tau,1},\ldots,\sS_{\tau,K_\tau}\},
$
on a common observation window $[0,T]$. For example, a task could
correspond to neuronal spike trains under one experimental condition, with the task set $\sD_\tau$ representing
repeated recordings under that condition.  Different tasks may correspond to different conditions, such as odor versus visual stimulation, and hence have different underlying firing intensities. Each task has its own unknown latent
intensity function
$\lambda_\tau:[0,T]\rightarrow[0,\infty)$, and the point patterns within the
task are modelled as conditionally independent realizations from this shared
intensity:
\begin{equation}
\sS_{\tau,k} \mid \lambda_\tau
\overset{\mathrm{iid}}{\sim}
\operatorname{PoissonProcess}(\lambda_\tau),
\qquad
k=1,\ldots,K_\tau.
\label{eq:replicated_pp}
\end{equation}
The intensity therefore varies across tasks but is shared across realizations
within a task. In other words, the $K_\tau$ point patterns provide repeated evidence about the
same latent function.

\begin{figure}[t]
    \centering
    \includegraphics[width=\textwidth]{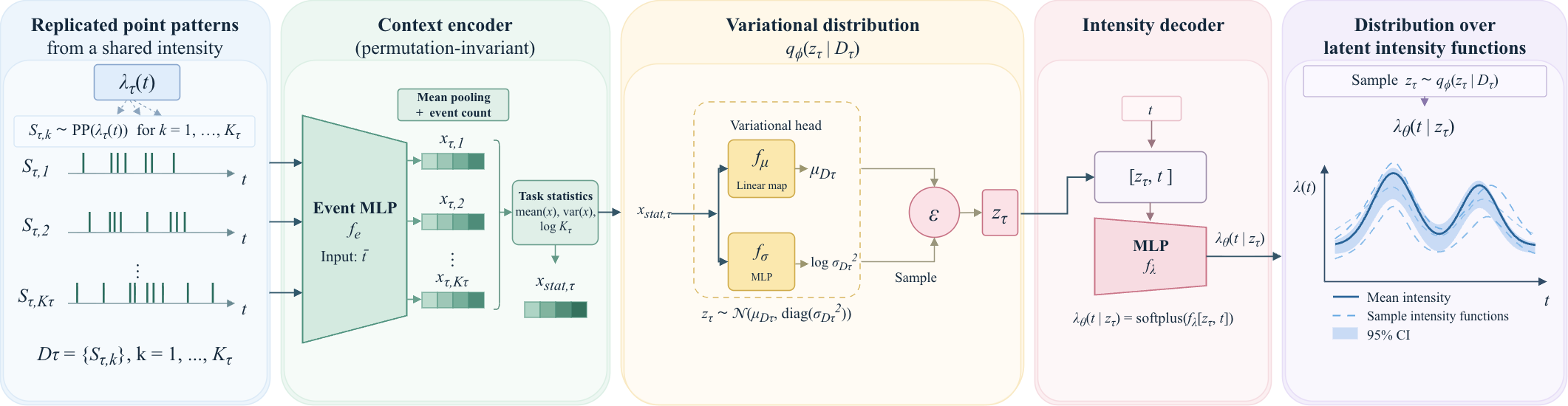}
    \caption{Architecture overview. Replicated point patterns parameterize a variational distribution whose samples are decoded into shared latent intensity functions.}
    \label{fig:model_architecture}
\end{figure}

Our goal is to design an inference procedure across a population of such tasks.
Given the observed point patterns $\sD_\tau$ of a new task, we seek to infer its shared latent intensity $\lambda_\tau$ while quantifying uncertainty about the function. Importantly,
the number of available realizations $K_\tau$ may differ across tasks and may
increase as further observations become available. The inference procedure
must therefore accept a variable-size, exchangeable set of point patterns and
combine the evidence they provide about the shared intensity. Drawing on
Neural Processes \citep{garnelo2018conditional,garnelo2018neural}, we learn a
single amortized inference rule which can be applied to newly observed tasks (possibly composed of a single point pattern) without per-task optimization.
Figure~\ref{fig:model_architecture} summarizes the framework.

Neural temporal point process models~\citep{du2016recurrent,omi2019fully,zuo2020transformer}
primarily model history-dependent intensities for event prediction. Recent work
extends these predictive models through meta-learning, but retains the same
prediction target. Meta Temporal Point Processes (Meta-TPP)~\citep{bae2023meta} formulate
each event sequence as a separate task and use a context set from that sequence
to predict its target event through an NP~\citep{garnelo2018conditional,garnelo2018neural}.
\citet{takimoto2024meta} encode short sequences into task representations
and use them to model intensities for predicting future events. FIM-PP~\citep{berghaus2026fimpp} uses variable-size sets of context sequences and a target history to estimate conditional intensity functions through in-context learning.

More closely related to our setting are methods that treat the intensity itself
as the object of inference. Deep Random Splines (DRS)~\citep{loaizaganem2019deep}
represent each channel's intensity as a random spline. Their
DRS-VAE uses separate recurrent encoders to process multiple point patterns.
It does not infer a single shared intensity function from exchangeable realizations.
Neural Diffusion Intensity Models (NDIM)~\citep{du2026neuraldiffusion}
model the intensity of a Cox process as a latent path governed by a neural stochastic
differential equation (SDE). This
formulation assumes that the intensity path is a Markov diffusion and performs amortized variational    
inference for each observed point pattern, rather than inferring a shared intensity.
LGCP-BayesFlow~\citep{wang2026lgcpbayesflow}
amortizes inference for the hyperparameters of Log-Gaussian Cox Process
(LGCP)~\citep{moller1998log} models from point patterns, rather than
the realized latent intensity function itself. Our focus is amortized distributional
inference for one latent intensity shared by a variable number of replicated
point patterns.
Table~\ref{tab:related} compares these methods with our approach.

Our contributions are as follows:\nopagebreak

\begin{itemize}
    \item We develop an amortized framework that maps replicated point patterns to a variational posterior distribution over their shared intensity, using the point process likelihood without ground-truth intensity supervision.

    \item We empirically demonstrate that the learned inference rule
continues to improve intensity recovery and decrease uncertainty as
additional realizations become available.
    \item We evaluate intensity recovery, uncertainty, and inference cost on synthetic data, and held-out model fit on bicycle-sharing data, demonstrating the utility of the learned inference rule without per-dataset optimization.
\end{itemize}

\begin{table}[t]
\caption{Comparisons of related point process methods.}
\label{tab:related}
\centering
\small
\renewcommand{\arraystretch}{1.18}
\setlength{\tabcolsep}{3pt}
\begin{tabular*}{\textwidth}{@{\extracolsep{\fill}}lcccc@{}}
\toprule
\textbf{Method}
& \shortstack{\textbf{Intensity}\\\textbf{inference}}
& \shortstack{\textbf{Intensity}\\\textbf{UQ}}
& \shortstack{\textbf{Replicated}\\\textbf{realizations}}
& \shortstack{\textbf{Amortized}\\\textbf{inference}} \\
\midrule
LGCP & $\checkmark$ & $\checkmark$ & $\checkmark$ & $\times$ \\
Meta-TPP & $\times$ & $\times$ & $\times$ & $\checkmark$ \\
FIM-PP & $\checkmark$ & $\times$ & $\checkmark$ & $\checkmark$ \\
LGCP-BayesFlow & $\times$ & $\times$ & $\checkmark$ & $\checkmark$ \\
DRS & $\checkmark$ & $\checkmark$ & $\times$ & $\checkmark$ \\
NDIM & $\checkmark$ & $\checkmark$ & $\times$ & $\checkmark$ \\
\midrule
\textbf{Ours} & $\boldsymbol{\checkmark}$ & $\boldsymbol{\checkmark}$ & $\boldsymbol{\checkmark}$ & $\boldsymbol{\checkmark}$ \\
\bottomrule
\end{tabular*}
\vspace{1pt}
\parbox{\textwidth}{\footnotesize\raggedright
}
\end{table}

\section{Preliminaries}

\subsection{Temporal Point Processes}

Temporal point processes~\citep{daley2003introduction} provide a probabilistic
framework for modeling event sequences occurring in continuous time. Let
$\sS=\{t_i\}_{i=1}^{N}$ denote an event sequence observed on a time window
$[0,T]$, with counting process $N(t)=\sum_{i=1}^{N}\mathbf 1\{t_i\leq t\}$ and
total count $N=N(T)$. Let $\mathcal H_t$ denote the history of events prior
to time $t$. Its conditional intensity is
$\lambda(t\mid\mathcal H_t)=\lim_{\Delta t\to 0^+}
\mathbb E[N(t+\Delta t)-N(t)\mid\mathcal H_t]/\Delta t$,
the instantaneous event rate given the observed history.
The log likelihood of the sequence $\sS$ is~\citep{daley2003introduction}:
\begin{equation}
    \log p(\sS\mid\lambda)
    =
    \sum_{i=1}^{N}
    \log \lambda(t_i\mid\mathcal H_{t_i})
    -
    \int_0^T
    \lambda(t\mid\mathcal H_t)\,\mathrm dt.
    \label{eq:tpp_likelihood}
\end{equation}
A Cox process~\citep{moller1998log} extends this formulation by treating the
intensity itself as a latent, non-negative random function $\lambda$. Conditional on a realization of
$\lambda$, the event sequence follows an inhomogeneous Poisson process,
$\sS\mid\lambda\sim\operatorname{PoissonProcess}(\lambda)$,
and its log likelihood is given by \Eqref{eq:tpp_likelihood} thus without history
dependence. A prominent instance is the
LGCP, which models the
log-intensity as a Gaussian process (GP),
$f\sim\operatorname{GP}(m,C)$ with $\lambda(t)=\exp\!\bigl(f(t)\bigr)$,
where $m$ and $C$ denote the mean and covariance functions. 

\subsection{Amortized Variational Inference}

Variational inference (VI)~\citep{blei2017variational} approximates an intractable
posterior $p(\rvz\mid x)$ within a tractable family by minimizing
Kullback--Leibler (KL) divergence. Amortized
VI~\citep{wu2020metaamortized}
replaces observation-specific optimization with an inference network
$q_\psi(\rvz\mid x)$ trained across observations $x\sim p_D$:
\begin{equation}
    \min_\psi\;
    \mathbb{E}_{x\sim p_D}
    \left[
        D_{\mathrm{KL}}\!\left(
            q_\psi(\rvz\mid x)
            \,\|\,
            p(\rvz\mid x)
        \right)
    \right].
    \label{eq:amortized_vi}
\end{equation}
Once trained, the network evaluates
$q_\psi(\rvz\mid x^\star)$ for an unseen observation $x^\star$ in a forward pass.

\subsection{Neural Processes}
\label{sec:np_preliminaries}

NPs~\citep{garnelo2018conditional,garnelo2018neural,dubois2020npf}
combine stochastic process modeling with amortized neural inference.
They are trained across a collection of tasks, where each task
consists of observations generated from an underlying function, while different
tasks correspond to different functions. By learning similarities across tasks, NPs can adapt to a previously unseen function from
a small number of observations in a forward pass.

Within each task, observations are partitioned into a \emph{context set}
and a \emph{target set},
\begin{equation}
    \sC=\{(x_i,y_i)\}_{i=1}^{N_C},
    \qquad
    \sT=\{(x_j,y_j)\}_{j=1}^{N_T},
    \label{eq:np_sets}
\end{equation}
where $\sX_T=\{x_j\}_{j=1}^{N_T}$ and $\sY_T=\{y_j\}_{j=1}^{N_T}$ denote the target
inputs and outputs. The context and target sets contain input--output pairs associated with the
same underlying function. Given $\sC$ and the
target inputs $\sX_T$,  an NP models the conditional predictive distribution
$p(\sY_T\mid\sX_T,\sC)$.

Latent NPs introduce a task-level latent variable $\rvz$ to capture uncertainty
about the underlying function~\citep{garnelo2018neural}. An
encoder maps the context set to a conditional latent distribution
$q_\phi(\rvz\mid\sC)$, while a decoder generates predictions for the target set conditioned on $\rvz$.
The resulting predictive distribution is
\begin{equation}
    p(\sY_T\mid\sX_T,\sC)
    =
    \int
    p(\sY_T\mid\sX_T,\rvz)\,
    q_\phi(\rvz\mid\sC)\,
    \mathrm d\rvz,
    \quad
    p(\sY_T\mid\sX_T,\rvz)
    =
    \prod_{j=1}^{N_T}
    p(y_j\mid x_j,\rvz).
    \label{eq:np_target_marginal}
\end{equation}

\section{Method}

\subsection{Problem Formulation}
\label{sec:problem}

We now extend the setup from Section~\ref{sec:intro} to a hierarchical model where each task $\tau\sim p(\tau)$
is characterized by an unknown nonnegative intensity function $\lambda_\tau:[0,T]\to[0,\infty)$,
and is observed through a dataset $\sD_\tau$ of $K_\tau\geq 1$ iid realizations:
\begin{equation}
    \sD_\tau=\{\sS_{\tau,k}\}_{k=1}^{K_\tau},
    \qquad
    \sS_{\tau,k}=\{t_{\tau,k,i}\}_{i=1}^{N_{\tau,k}} \mid \lambda_\tau 
    \overset{\mathrm{iid}}{\sim}
    \operatorname{PoissonProcess}(\lambda_\tau),
    \label{eq:cox_task}
\end{equation}
with $t_{\tau,k,i}\in [0,T]$ denoting the $i$-th event time in the $k$-th sequence with event count $N_{\tau,k}$.

Given $M$ training tasks $\{\sD_\tau\}_{\tau=1}^{M}$, our goal is to learn an amortized
inference rule $\mathcal A$ that maps each task to a variational
distribution over its shared intensity function,
$\mathcal A:\sD_\tau\mapsto\mathcal{Q}(\lambda_\tau\mid\sD_\tau)$.
The rule $\mathcal A$ is shared across tasks and accepts a variable number
of replications $K_\tau$. Consequently, for a new task $\tau^\star$,
inference is performed in a forward pass.

\subsection{Model Overview}
\label{sec:model_overview}

We represent each task by a latent variable
$\rvz_\tau\in\R^{d_z}$ that captures information about its unknown intensity
$\lambda_\tau$. A decoder parameterized by $\theta$ maps $\rvz_\tau$
to the intensity
$\lambda_\tau(t)=\lambda_\theta(t\mid\rvz_\tau)$.

Since conditionally on the shared intensity $\lambda_\tau$, the $K_\tau$
realizations are independent,  using \Eqref{eq:tpp_likelihood}, their joint likelihood is
\begin{equation}
    p_\theta(\sD_\tau\mid\rvz_\tau)
    =
    \exp\!\left(
        -K_\tau\int_0^T
        \lambda_\theta(t\mid\rvz_\tau)\,\mathrm{d}t
    \right)
    \prod_{k=1}^{K_\tau}
        \prod_{i=1}^{N_{\tau,k}}
        \lambda_\theta(t_{\tau,k,i}\mid\rvz_\tau).
    \label{eq:task_point_process_likelihood}
\end{equation}
Given the observed realizations $\sD_\tau$, a context encoder parameterized by
$\phi$ defines a learned latent distribution
$q_\phi(\rvz_\tau\mid\sD_\tau)$.
The encoder--decoder combination induces a variational distribution over the shared intensity of task $\tau$, $q_{\phi,\theta}(\lambda_\tau\mid\sD_\tau)$, as defined in \Eqref{eq:intensity_variational_posterior}.
We train the context encoder and decoder using the context--target conditional objective in
Section~\ref{sec:training_objective}. We next describe the context encoder, decoder, and training objective.

\subsection{Context Encoder}
\label{sec:context_encoder}

The context encoder maps the observed realizations $\sD_\tau$ to the
latent distribution $q_\phi(\rvz_\tau\mid\sD_\tau)$ via
aggregation over events and realizations.

Each event $t_{\tau,k,i}$ is standardized as $\bar t_{\tau,k,i}\in\R$ using training statistics
and embedded via an MLP $f_e$:
\begin{equation}
\vh_{\tau,k,i}
=
f_e(\bar t_{\tau,k,i})
\in\R^{d_h}.
\label{eq:event_encoder}
\end{equation}
We perform mean pooling across events within each realization $\sS_{\tau,k}$ to obtain the realization embedding $\vu_{\tau,k}$, and form the augmented realization vector $\vx_{\tau,k}$:
\begin{equation}
\vu_{\tau,k}
=
\frac{1}{N_{\tau,k}}
\sum_{i=1}^{N_{\tau,k}}
\vh_{\tau,k,i},
\qquad
\vx_{\tau,k} = [\vu_{\tau,k}^\top, \log(1+N_{\tau,k})]^\top \in \R^{d_h+1},
\label{eq:realization_encoder}
\end{equation}
Mean pooling ensures permutation invariance within each realization, while
$\log(1+N_{\tau,k})$ retains information about its event count. 
To quantify variation across realizations and adapt the size of uncertainty sets to $K_\tau$, we augment
the realization vectors with information about the marginal variance across realizations and the number of replicated sequences:
\begin{equation}
    \vx_{\mathrm{stat},\tau}
    =
    \left[
        \operatorname{mean}_{k}(\vx_{\tau,k}),\;
        \operatorname{var}_{k}(\vx_{\tau,k}),\;
        \log K_\tau
    \right] \in \R^{2(d_h+1)+1},
    \label{eq:variance_features}
\end{equation}
where the variance is computed coordinate-wise across realizations.
The mean and log-variance heads both operate on $\vx_{\mathrm{stat},\tau}$:
\begin{equation}
    \vmu_{\sD_\tau} = f_\mu(\vx_{\mathrm{stat},\tau}),
    \qquad
    \bm\sigma_{\sD_\tau}^2 = \exp\!\left(f_\sigma(\vx_{\mathrm{stat},\tau})\right),
\label{eq:latent_heads}
\end{equation}
Here, $f_\mu$ is a linear map and $f_\sigma$ is an MLP parameterizing the latent mean and log-variance, respectively.
These heads yield the diagonal Gaussian variational posterior as in a Variational Autoencoder (VAE)~\citep{kingma2014auto},
\begin{equation}
q_\phi(\rvz_\tau\mid\sD_\tau)
=
\mathcal N\!\left(
\vmu_{\sD_\tau},
\operatorname{diag}
(\bm\sigma_{\sD_\tau}^{2})
\right).
\label{eq:context_posterior}
\end{equation}
Architecture and standardization details are provided in
Appendix~\ref{app:context_encoder_architecture}.

\subsection{Intensity Decoder}
\label{sec:intensity_decoder}

The decoder maps the latent variable $\rvz_\tau$ and query time
$t$ to a continuous-time intensity using an MLP $f_\lambda$ with a
softplus~\citep{zhang2020selfattentive,sulem2024bayesian,zuo2020transformer} output:
\begin{equation}
    \lambda_\tau(t)
    =
    \lambda_\theta(t\mid\rvz_\tau)
    =
    \operatorname{softplus}\!\left(
        f_\lambda[\rvz_\tau, t]
    \right).
    \label{eq:intensity_decoder}
\end{equation}
The latent distribution therefore induces a distribution over intensity
functions,
\begin{equation}
    q_{\phi,\theta}(\lambda_\tau\mid\sD_\tau)
    =
    \int
    p_\theta(\lambda_\tau\mid\rvz_\tau)
    q_\phi(\rvz_\tau\mid\sD_\tau)
    \,\mathrm d\rvz_\tau.
    \label{eq:intensity_variational_posterior}
\end{equation}
Decoder architecture is detailed in
Appendix~\ref{app:intensity_decoder_architecture}.

\subsection{Variational Inference}
\label{sec:training_objective}

For comparison, standard VI for a latent variable model
with a fixed prior $p(\rvz_\tau)=\mathcal N(\vzero,\mI)$ optimizes an evidence lower bound (ELBO) over the
complete dataset $\sD_\tau$,
\begin{equation}
    \log p_\theta(\sD_\tau)
    \geq
    \mathbb E_{q_\phi(\rvz_\tau\mid\sD_\tau)}
    \!\left[
        \log p_\theta(\sD_\tau\mid\rvz_\tau)
    \right]
    -
    D_{\mathrm{KL}}\!\left[
        q_\phi(\rvz_\tau\mid\sD_\tau)
        \,\middle\|\,
        p(\rvz_\tau)
    \right].
    \label{eq:standard_joint_elbo}
\end{equation}
We train a conditional model to predict held-out realizations from an
observed context. Following the context--target construction of
NPs~\citep{garnelo2018neural}, we partition each training task into disjoint
context and target sets,
\begin{equation}
    \sC_\tau
    =
    \{\sS_{\tau,k}\}_{k=1}^{K_C},
    \qquad
    \sT_\tau
    =
    \{\sS_{\tau,k}\}_{k=K_C+1}^{K_\tau},
    \qquad
    \sC_\tau\cup\sT_\tau=\sD_\tau,
    \qquad
    K_\tau=K_C+K_T.
    \label{eq:context_target_partition}
\end{equation}
While inference at test time accommodates any $K\geq1$, training requires
$K_\tau\geq2$ to ensure non-empty context and target subsets.

Given context $\sC_\tau$, the encoder defines a context-conditioned latent
distribution $q_\phi(\rvz_\tau\mid\sC_\tau)$. Following the NP construction,
we use this distribution as a context-conditioned prior, inducing the
conditional model
\begin{equation}
    \widetilde p_{\phi,\theta}(\sT_\tau\mid\sC_\tau)
    =
    \int
    p_\theta(\sT_\tau\mid\rvz_\tau)
    q_\phi(\rvz_\tau\mid\sC_\tau)
    \,\mathrm d\rvz_\tau.
    \label{eq:conditional_np_model}
\end{equation}
The latent posterior under this conditional model is
\begin{equation}
    \widetilde p_{\phi,\theta}
    (\rvz_\tau\mid\sT_\tau,\sC_\tau)
    =
    \frac{
        p_\theta(\sT_\tau\mid\rvz_\tau)
        q_\phi(\rvz_\tau\mid\sC_\tau)
    }{
        \widetilde p_{\phi,\theta}(\sT_\tau\mid\sC_\tau)
    }.
    \label{eq:conditional_latent_posterior}
\end{equation}
During training, $q_\phi(\rvz_\tau\mid\sC_\tau\cup\sT_\tau)$ serves as a
variational approximation to this posterior, giving the conditional ELBO
\begin{equation}
    \log
    \widetilde p_{\phi,\theta}(\sT_\tau\mid\sC_\tau)
    \geq
    \mathbb E_{q_\phi(\rvz_\tau\mid\sC_\tau\cup\sT_\tau)}
    \!\left[
        \log p_\theta(\sT_\tau\mid\rvz_\tau)
    \right]
    -
    D_{\mathrm{KL}}\!\left[
        q_\phi(\rvz_\tau\mid\sC_\tau\cup\sT_\tau)
        \,\middle\|\,
        q_\phi(\rvz_\tau\mid\sC_\tau)
    \right].
    \label{eq:training_objective}
\end{equation}

Conditional on $\rvz_\tau$, the target realizations are independent, with
conditional log likelihood
\begin{equation}
    \log p_\theta(\sT_\tau\mid\rvz_\tau)
    =
    \sum_{\sS\in\sT_\tau}
    \left[
        \sum_{t_i\in\sS}
        \log \lambda_\theta(t_i\mid\rvz_\tau)
        -
        \int_0^T
        \lambda_\theta(t\mid\rvz_\tau)\,\mathrm dt
    \right].
    \label{eq:target_conditional_loglik}
\end{equation}

The encoder, and hence the resulting conditional model, 
is permutation invariant with respect to the ordering of replicated point patterns.
A derivation of the conditional objective is provided in
Appendix~\ref{app:theory_and_training}.

\section{Experiments}
We test whether amortized inference can recover shared latent intensities without per-task fitting, and whether recovery and uncertainty adapt as additional realizations become available.

We evaluate the proposed method along three dimensions: (i) recovery and uncertainty quantification for latent intensity functions, (ii) adaptation as additional replicated point patterns become available, and (iii) computational efficiency relative to per-task inference. We first study these properties in a controlled synthetic setting, where the latent intensity is known, and then evaluate held-out model fit on real bicycle-sharing data. We approximate the likelihood integral over $[0,T]$ using the trapezoidal rule on $G=300$ equally spaced points and estimate the expectation in \Eqref{eq:training_objective} with $m=3$ reparameterized latent samples per task.

\subsection{Synthetic Experiments}
\label{sec:synthetic_experiments}

Synthetic experiments allow us to evaluate inference against known latent
intensity functions. We assess whether the proposed amortized inference
procedure accurately recovers these intensities and quantifies uncertainty,
and whether both intensity recovery and uncertainty adapt as additional
replicated point patterns become available. We further examine whether this
evidence accumulation extends to numbers of realizations beyond those
encountered during training.

\subsubsection{Experimental Setup}
\label{sec:synthetic_setup}

\noindent\textbf{Data generation and protocol.}
For each task $\tau$, we independently draw a latent function on $[0,10]$
\begin{equation}
    f_\tau \sim
    \operatorname{GP}\!\left(
        0,
        C_{\mathrm{RBF}}
    \right),
    \qquad
    \lambda_\tau(t)
    =
    20\operatorname{softplus}(f_\tau(t)),
    \label{eq:synthetic_intensity}
\end{equation}
where
$C_{\mathrm{RBF}}(t,t')
=\sigma_f^2\exp(-(t-t')^2/(2\ell^2))$
with signal variance $\sigma_f^2=1$ and lengthscale $\ell=2$. Conditionally on $\lambda_\tau$, the realizations within each task are generated independently from the corresponding Poisson process. Thus, each synthetic task has a distinct latent intensity, while its replicated point patterns provide repeated observations of that same intensity.

We use 4,500 training, 500 validation, and 500 test tasks, denoted by
$\mathcal D_{\mathrm{train}}$, $\mathcal D_{\mathrm{val}}$, and
$\mathcal D_{\mathrm{test}}$, respectively.
Neither training nor checkpoint selection uses ground-truth intensity labels.

For our method, we select checkpoints based on held-out model fit: how well the
inferred intensity distribution explains realizations excluded from its
context. For each task $\sD_\tau$, we hold out each realization
$\sS_{\tau,h}$ in turn and use the remaining realizations
$\sD_{\tau,-h}$ as context. For
$s\in\{\mathrm{val},\mathrm{test}\}$, we define the held-out log
likelihood (HLL) as
\begin{equation}
    \operatorname{HLL}_s
    =
    \frac{1}{|\mathcal D_s|}
    \sum_{\sD_\tau\in\mathcal D_s}
    \frac{1}{|\sD_\tau|}
    \sum_{h=1}^{|\sD_\tau|}
    \log
    \mathbb E_{q_{\phi,\theta}(\lambda_\tau\mid\sD_{\tau,-h})}
    \left[
        p(\sS_{\tau,h}\mid\lambda_\tau)
    \right].
    \label{eq:heldout_hll}
\end{equation}
The expectation is approximated using 300 latent samples. After training, we compute the HLL of each saved 
checkpoint on the validation tasks and select 
the checkpoint with the highest validation HLL separately for each 
training seed. Further training and selection details are provided 
in Appendix~\ref{app:synthetic_selection}.

To study the effect of the number of training realizations, 
we fix $K_\tau=K_{\mathrm{train}}$ for all training tasks 
and sample the context size as $K_C \sim \frac{1}{2}\mathrm{Unif}\{\lceil K_{\mathrm{train}}/2\rceil,\ldots,K_{\mathrm{train}}-1\}+\frac{1}{2}\mathrm{Unif}\{1,\ldots,K_{\mathrm{train}}-1\}$, 
with the target size given by $K_T = K_{\mathrm{train}} - K_C$, 
where $\mathrm{Unif}$ denotes the discrete uniform distribution 
over the specified set. This sampling scheme favors $K_C\geq K_T$ on average 
to limit the information gap between the context and full variational distributions.

We train models with $K_{\mathrm{train}}\in\{4,8,16\}$. At test time, we progressively increase the number $K_{\mathrm{test}}$ of realizations supplied to the encoder, using nested subsets from the same task. This allows us to assess whether a fixed trained model continues to accumulate evidence as additional realizations become available, including when $K_{\mathrm{test}}> K_{\mathrm{train}}$. All configurations are trained for 20,000 updates with four random seeds. Dataset statistics and further experimental details are provided in Appendices~\ref{app:experimental_configuration} and~\ref{app:synthetic_selection}.

\subsubsection{Baselines and Metrics}
\label{sec:synthetic_baselines_metrics}

\noindent\textbf{Baselines.}
To evaluate recovery and uncertainty, we compare against two classical nonparametric intensity estimators, a Bayesian task-specific estimator and an amortized neural method:
(i)~\textbf{PSTH}~\citep{kass2003statistical}, a peristimulus time histogram with bin width selected on the validation set;
(ii)~\textbf{KDE}~\citep{diggle1985kernel}, a kernel estimator with bandwidth selected on the validation set;
(iii)~\textbf{LGCP}, a GP intensity model with an exponential link fitted to each task using sparse variational inference; and
(iv)~\textbf{FIM-PP}~\citep{berghaus2026fimpp}, a pretrained model designed for conditional intensity prediction.
LGCP additionally provides uncertainty estimates, whereas PSTH, KDE, and FIM-PP provide point estimates only. 

\noindent\textbf{Metrics.}
We report intensity \textbf{RMSE} using the variational posterior mean, pointwise 95\% credible interval (CI) coverage (\textbf{C95}), mean 95\% CI width (\textbf{IW95}), and per-task runtime. C95 and IW95 apply only to methods with uncertainty estimates. For our method and LGCP, RMSE and interval metrics are estimated using 300 posterior intensity samples. 
Details are given in Appendices~\ref{app:synthetic_metrics} and~\ref{app:synthetic_baselines_runtime}.

\subsubsection{Results}
\label{sec:synthetic_results}

\noindent\textbf{Recovery and uncertainty.}
Table~\ref{tab:synthetic_baselines} shows that our method achieves the lowest RMSE (1.55), compared with 1.86 for LGCP and 1.91 for KDE. Among the methods that quantify uncertainty, our method attains empirical 95\% coverage of \(93.28\%\), compared with \(92.70\%\) for LGCP, while producing narrower credible intervals (5.73 versus 6.67). Thus, the reduction in interval width is accompanied by comparable empirical coverage, although both methods remain below the nominal \(95\%\) level. 

\noindent\textbf{Accuracy--efficiency trade-off.}
Once trained, our method performs inference for a new task without task-specific optimization, requiring 7.53 ms per task compared with 6.31 s for LGCP. The classical PSTH and KDE estimators are faster, requiring 0.04 ms and 0.75 ms, respectively, but yield higher RMSE and do not provide uncertainty estimates. Our method therefore combines low inference cost with accurate intensity recovery and uncertainty quantification.

\begin{table}[htbp]
    \centering
    \setlength{\belowcaptionskip}{4pt}
    \caption{Synthetic results on 500 test tasks with $K_{\mathrm{test}}=4$. C95 is reported as a percentage (\%). For learned models, $\pm$ denotes standard deviation across four training seeds on the same test set. Classical baseline metrics are deterministic.}
    \label{tab:synthetic_baselines}
    \small
    \setlength{\tabcolsep}{11pt}
    \begin{tabular}{lcccc}
        \toprule
        & \textbf{Recovery} & \multicolumn{2}{c}{\textbf{Uncertainty}} & \textbf{Efficiency} \\
        \cmidrule(lr){2-2}\cmidrule(lr){3-4}\cmidrule(lr){5-5}
        \textbf{Method} & \textbf{RMSE} $\downarrow$ & \textbf{C95 (\%)} & \textbf{IW95}  & \textbf{Time (ms)} $\downarrow$ \\
        \midrule
        FIM-PP           & 14.19 & --- & --- & 5.81 \\
        PSTH             & 2.45 & --- & --- & 0.04 \\
        KDE              & 1.91 & --- & --- & 0.75 \\
        \midrule
        LGCP             & $1.86_{\pm 0.01}$ & $92.70_{\pm 0.10}$ & $6.67_{\pm 0.01}$ & $6{,}311.76_{\pm 36.06}$ \\
        \midrule
        \textbf{Ours ($K_{\mathrm{train}}=4$)} & $\mathbf{1.55}_{\pm 0.01}$ & ${93.28}_{\pm 0.18}$ & ${5.73}_{\pm 0.00}$ & ${7.53_{\pm 0.44}}$ \\
        \bottomrule
    \end{tabular}
\end{table}

\begin{figure}[!t]
    \centering
    \includegraphics[width=\textwidth]{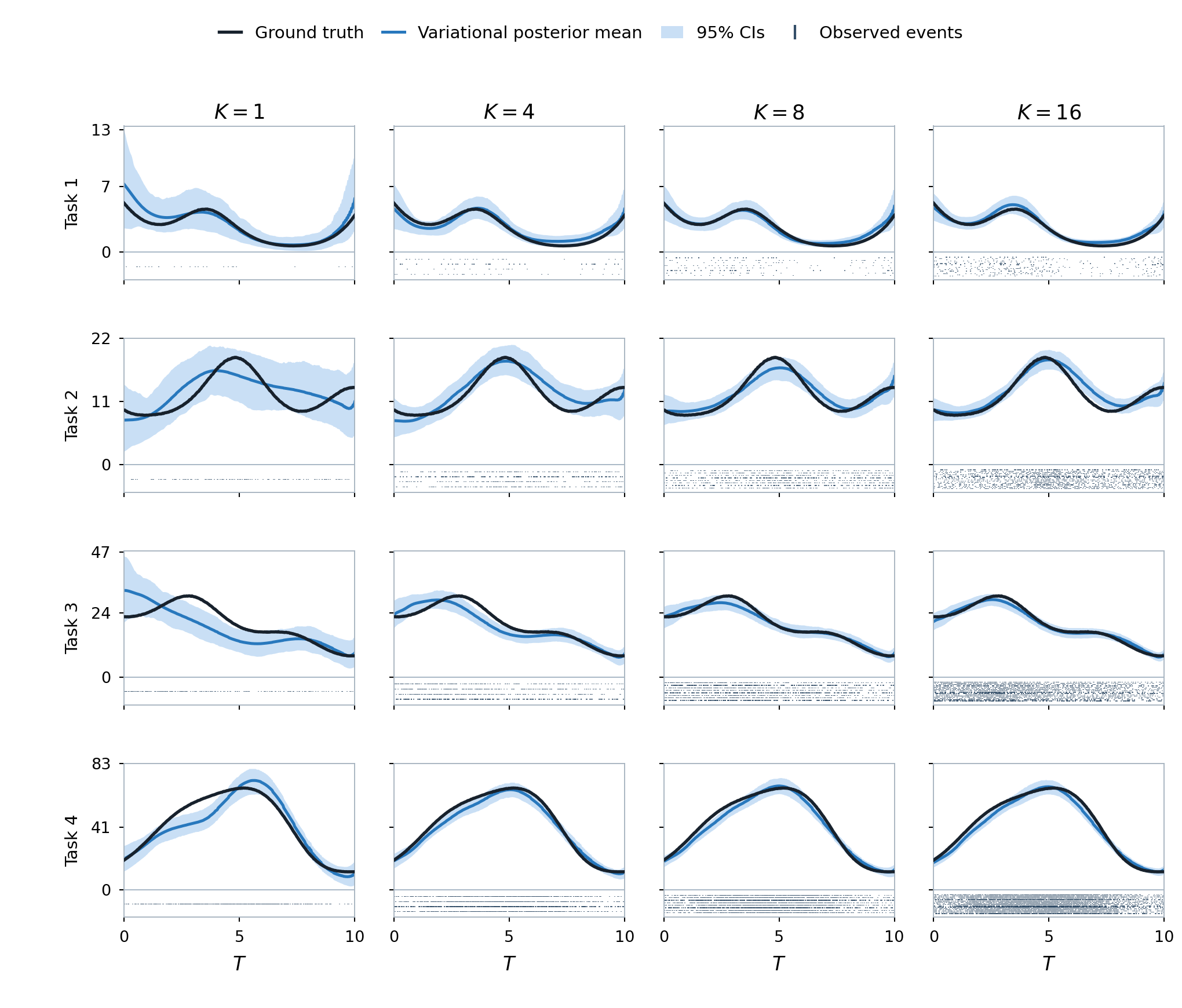}
    \caption{Adaptation to additional realizations for the model trained with $K_{\mathrm{train}}=8$. Four test tasks are shown as $K_{\mathrm{test}}\in\{1,4,8,16\}$ realizations become available. Across these examples, additional realizations generally improve intensity recovery and yield narrower CIs.}
    \label{fig:intensity_samples}
\end{figure}

\noindent\textbf{Evidence accumulation.}
Table~\ref{tab:synthetic_cross_k} shows that the model trained with
$K_{\mathrm{train}}=4$ benefits from additional realizations. Increasing
$K_{\mathrm{test}}$ from 4 to 64 reduces RMSE from 1.55 to 0.82 and the mean
95\% interval width from 5.73 to 3.74, while coverage remains similar
(93.28\% and 93.16\%, respectively). Figure~\ref{fig:intensity_samples}
illustrates this behavior: as additional realizations from the same task are
observed, the posterior mean approaches the true intensity and the credible
intervals become narrower. Results for $K_{\mathrm{train}}\in\{8,16\}$ are
reported in Appendix~\ref{app:additional_ktrain_results}.
Additional out-of-distribution evaluations are reported in Appendix~\ref{app:synthetic_ood}.

\begin{table}[!htbp]
    \centering
    \caption{Evidence accumulation for the model trained with $K_{\mathrm{train}}=4$. Results are averaged over four seeds.}
    \label{tab:synthetic_cross_k}
    \small
    \begin{tabular}{cccc}
        \toprule
        $K_{\mathrm{test}}$ & \textbf{RMSE} $\downarrow$ & \textbf{C95 (\%)} & \textbf{IW95} \\
        \midrule
        1  & $2.68_{\pm 0.01}$ & $92.21_{\pm 0.05}$ & $9.36_{\pm 0.02}$ \\
        2  & $1.99_{\pm 0.00}$ & $93.18_{\pm 0.09}$ & $7.20_{\pm 0.02}$ \\
        4  & $1.55_{\pm 0.01}$ & $93.28_{\pm 0.18}$ & $5.73_{\pm 0.00}$ \\
        8  & $1.23_{\pm 0.01}$ & $93.78_{\pm 0.25}$ & $4.78_{\pm 0.01}$ \\
        16 & $1.00_{\pm 0.01}$ & $94.32_{\pm 0.28}$ & $4.23_{\pm 0.01}$ \\
        32 & $0.88_{\pm 0.01}$ & $93.84_{\pm 0.21}$ & $3.91_{\pm 0.02}$ \\
        64 & $0.82_{\pm 0.02}$ & $93.16_{\pm 0.30}$ & $3.74_{\pm 0.03}$ \\
        \bottomrule
    \end{tabular}
\end{table}

\subsection{Real-data experiments}
\label{sec:real_data}

We use the dataset of Divvy bicycle-share trip records from Chicago, USA, spanning
2013--2019.\footnote{City of Chicago, \emph{Divvy Trips}:
\url{https://data.cityofchicago.org/d/fg6s-gzvg}.}
Each trip start is treated as an event, and a point pattern consists of the trip start times
observed at one station on a given day. To construct replicated observations, we group four days matching in station, year, month, and weekday into a task. Thus, each task contains $K=4$ point patterns that are treated as approximate realizations of a shared daily intensity. Calendar matching and additional count-homogeneity screening are used to make this shared intensity assumption more plausible. Event times are rescaled to map each day to $[0,10]$.

After screening and subsampling, we obtain train/validation/test splits
containing 5,387/674/674 tasks from 182/21/21 stations (224 total).
Stations are assigned disjointly across the three splits, 
so that all test tasks correspond to stations that were unseen during training.
Further preprocessing and screening details are given in Appendix~\ref{app:real_data_construction}.

We report HLL as defined in \Eqref{eq:heldout_hll}.
LGCP follows the same evaluation protocol, whereas PSTH, KDE, and
FIM-PP use their point intensity estimates.
Further details are provided in
Appendices~\ref{app:synthetic_selection}
and~\ref{app:real_data_baselines}.

We train our model on $\mathcal D_{\mathrm{train}}$ for 20,000 updates. 
After training, we evaluate all saved checkpoints on $\mathcal D_{\mathrm{val}}$ using the same four-fold HLL with 300 latent samples and select the checkpoint with the highest validation HLL separately for each training seed. The validation set is also used to select the PSTH bin width and KDE bandwidth. Training details are provided in Appendix~\ref{app:real_data_training}. Table~\ref{tab:real_data_comparison} reports the test-set results.

On tasks from previously unseen stations, our model infers a
distribution over the shared intensity from three replicated point
patterns. It achieves the highest mean
HLL among the evaluated methods
(Table~\ref{tab:real_data_comparison}), outperforming LGCP, which requires
per-task optimization. Mean computation time is 26.82 ms
per task, compared with 24.4 seconds for LGCP. On unseen stations, the
learned inference rule therefore combines strong held-out model fit with
substantially lower computational cost.

\begin{figure}[H]
    \centering
    \includegraphics[width=0.8\textwidth]{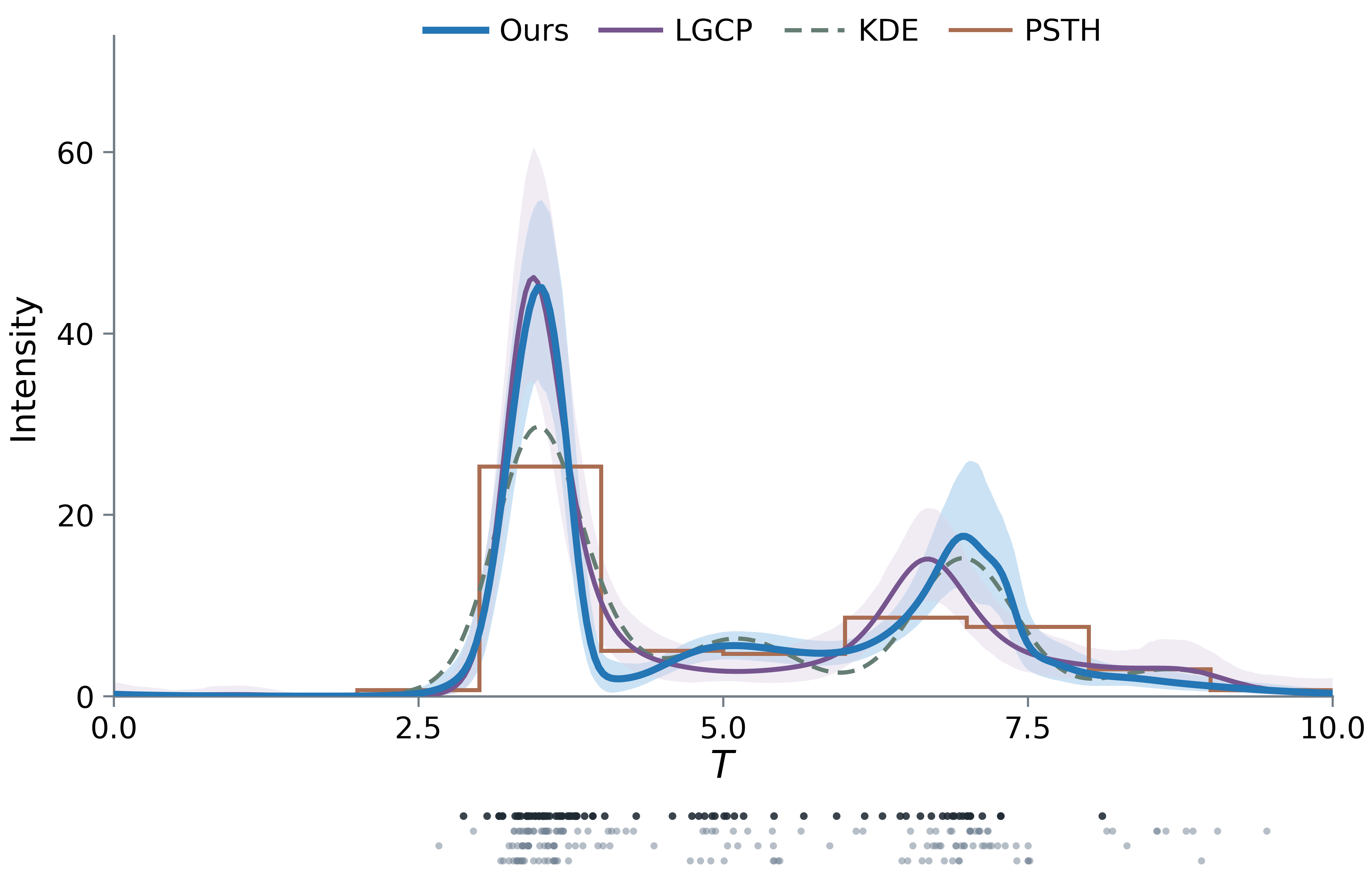}
    \caption{
Held-out model fit for a test task. All estimates use only the three context realizations shown in the
lower, light rows. The upper, dark row shows the held-out realization
and is used only for evaluation. Solid curves show the inferred mean
intensities for Ours and LGCP, with 95\% CIs; the dashed
and step curves show the KDE and PSTH estimates, respectively.
    }
    \label{fig:real_data_intensities}
\end{figure}

Figure~\ref{fig:real_data_intensities} provides a qualitative held-out
predictive check. All curves are inferred using only the three context
realizations shown in the lower, light rows; the upper, dark row is a
fourth realization used only for evaluation. All methods recover the two
main periods of elevated activity, around $t\approx3.5$ and $t\approx7$.
Our inferred mean preserves a low-intensity region between the peaks and
produces a more localized second peak, where the held-out events also
concentrate.
\begin{table}[H]
    \centering
    \setlength{\belowcaptionskip}{4pt}
    \caption{Divvy results (held-out log likelihood and time) on 674 test tasks. $\pm$ denotes standard deviation across four seeds for Ours and LGCP.}
    \label{tab:real_data_comparison}
    \small
    \setlength{\tabcolsep}{11pt}
    \begin{tabular}{lcc}
        \toprule
        \textbf{Method} & \textbf{HLL} $\uparrow$ & \textbf{Time (ms)} $\downarrow$ \\
        \midrule
        FIM-PP           & 59.53 & 76.03 \\
        PSTH             & 102.73 & \textbf{0.19} \\
        KDE              & 107.10 & 10.41 \\
        \midrule
        LGCP             & $108.80_{\pm 0.01}$ & $24{,}407.47_{\pm 76.83}$ \\
        \midrule
        \textbf{Ours}    & $\mathbf{110.99}_{\pm 0.03}$ & $26.82_{\pm 0.13}$ \\
        \bottomrule
    \end{tabular}
\end{table}

\section{Conclusion}
We presented an amortized variational framework for inferring a shared
latent intensity from replicated point patterns using the point process
likelihood. The learned inference rule improves intensity recovery and
narrows credible intervals as additional realizations become available,
including beyond the training replication count, with broadly stable
empirical coverage in this extrapolation regime. On bicycle-sharing data,
it achieves the highest mean held-out log likelihood among the evaluated
methods, while requiring substantially less inference time than LGCP.
Together, these results show that a single trained model can refine
intensity estimates and their uncertainty as evidence accumulates,
without per-task optimization. Quantifying and reducing the 
amortization gap is an important direction for future work.
\section*{Acknowledgements}
We acknowledge access to Piz Daint or Alps at the Swiss National Supercomputing Centre, Switzerland under the Universit\`a della Svizzera italiana's share with the project ID u0.

XM has received funding from the European Union's Horizon Europe research and innovation programme under the Marie Sklodowska-Curie grant agreement 101151781 and from the University of Cyprus under the startup grant programme.

\subsection*{Reproducibility statement}

We provide code and data for our proposed method in our \href{https://github.com/Wenqing-Liu-1010/Amortized-Variational-Inference-of-Latent-Intensities-from-Replicated-Point-Patterns}{GitHub repository}. The repository includes scripts for model training and evaluation, the configurations and random seeds used for the reported experiments with our method, as well as scripts for producing its main evaluation results. We also provide the trained checkpoints used for evaluation.
\clearpage
\bibliography{references}
\bibliographystyle{plainnat}

\appendix
\section{Encoder and Decoder Architectures}
\label{app:model_architecture}

\subsection{Context Encoder}
\label{app:context_encoder_architecture}

Let $\mathcal{T}_{\mathrm{train}}$ be the collection of event times in the
training tasks. The standardized event time used in
\Eqref{eq:event_encoder} is
\begin{equation}
    m_t=\frac{1}{|\mathcal{T}_{\mathrm{train}}|}\sum_{t\in\mathcal{T}_{\mathrm{train}}}t,
    \qquad
    s_t=\sqrt{\frac{1}{|\mathcal{T}_{\mathrm{train}}|}
    \sum_{t\in\mathcal{T}_{\mathrm{train}}}(t-m_t)^2},
    \qquad
    \bar t_{\tau,k,i}=\frac{t_{\tau,k,i}-m_t}{s_t}.
    \label{eq:event_standardization}
\end{equation}
The statistics $m_t$ and $s_t$ are fixed for validation and test tasks.

For compactness, define
\begin{equation*}
    c_{\tau,k}=\log(1+N_{\tau,k}),
    \qquad
    \bar{\vx}_\tau=\frac{1}{K_\tau}\sum_{k=1}^{K_\tau}\vx_{\tau,k},
\end{equation*}
\begin{equation*}
    \bm v_\tau=\frac{1}{K_\tau}\sum_{k=1}^{K_\tau}
    (\vx_{\tau,k}-\bar{\vx}_\tau)^{\odot 2},
    \qquad
    \kappa_\tau=\log K_\tau.
\end{equation*}
Table~\ref{tab:context_encoder_architecture} presents the encoder as a sequence of modules.

\begin{table}[htbp]
    \centering
    \caption{Context encoder architecture.}
    \label{tab:context_encoder_architecture}
    \footnotesize
    \setlength{\tabcolsep}{3pt}
    \begin{tabular*}{\textwidth}{@{\extracolsep{\fill}}llll@{}}
        \toprule
        Module & Input & Transformation & Output \\
        \midrule
        Event encoder $f_e$
        & $\bar t_{\tau,k,i}\in\R$
        & $\operatorname{MLP}_{\mathrm{SiLU}}(1,512,512,512,256)$
        & $\vh_{\tau,k,i}\in\R^{256}$ \\
        \addlinespace
        Realization representation
        & $\{\vh_{\tau,k,i}\}_{i=1}^{N_{\tau,k}}$
        & $\vu_{\tau,k}=N_{\tau,k}^{-1}\sum_i\vh_{\tau,k,i}$;
          $\vx_{\tau,k}=[\vu_{\tau,k}^{\top},c_{\tau,k}]^{\top}$
        & $\vx_{\tau,k}\in\R^{257}$ \\
        \addlinespace
        Task statistics
        & $\{\vx_{\tau,k}\}_{k=1}^{K_\tau}$
        & $\vx_{\mathrm{stat},\tau}
          =[\bar{\vx}_\tau^{\top},\bm v_\tau^{\top},\kappa_\tau]^{\top}$
        & $\vx_{\mathrm{stat},\tau}\in\R^{515}$ \\
        \addlinespace
        Mean head $f_\mu$
        & $\vx_{\mathrm{stat},\tau}\in\R^{515}$
        & $\operatorname{Linear}(515,256)$
        & $\vmu_{\sD_\tau}\in\R^{256}$ \\
        \addlinespace
        Log-variance head $f_\sigma$
        & $\vx_{\mathrm{stat},\tau}\in\R^{515}$
        & $\operatorname{MLP}_{\mathrm{SiLU}}(515,128,256)$
        & $\bm\ell_{\sD_\tau}\in\R^{256}$ \\
        \bottomrule
    \end{tabular*}
\end{table}

These two heads parameterize the diagonal Gaussian
\begin{equation}
q_\phi(\rvz_\tau\mid\sD_\tau)
=
\mathcal N\!\left(
\vmu_{\sD_\tau},
\operatorname{diag}(\bm\sigma_{\sD_\tau}^2)
\right),
\qquad
\bm\sigma_{\sD_\tau}^2
=
\exp(\bm\ell_{\sD_\tau}).
\end{equation}
For numerical stability, the raw output of the log-variance head is mapped
element-wise to $(-6,3)$. We then draw a differentiable
sample using the reparameterization trick:
\begin{equation}
\rvz_\tau
=
\vmu_{\sD_\tau}
+
\exp\!\left(\tfrac12\bm\ell_{\sD_\tau}\right)
\odot\bm\epsilon,
\qquad
\bm\epsilon\sim\mathcal N(\vzero,\mI).
\label{eq:reparameterization}
\end{equation}

\subsection{Intensity Decoder}
\label{app:intensity_decoder_architecture}

The decoder $f_\lambda$ concatenates a latent sample
$\rvz_\tau\in\R^{256}$ with a query time $t$. This input passes through three hidden layers of width
$512$ with SiLU activations, followed by a linear output layer and a softplus
transformation that produces the corresponding intensity
$\lambda_\theta(t\mid\rvz_\tau)$.

\section{Training Objective}
\label{app:theory_and_training}

This section distinguishes three related ELBOs. The first two belong to a
latent-variable model with Gaussian prior
$p(\rvz_\tau)=\mathcal N(\mathbf 0,\mI)$. The third is the objective used to
train our learned conditional model.

\noindent\textbf{1. Gaussian-prior ELBO.}
Combining the Gaussian prior with the likelihood
$p_\theta(\sD_\tau\mid\rvz_\tau)$ gives the standard ELBO in
\Eqref{eq:standard_joint_elbo}:
\begin{align}
    \log p_\theta(\sD_\tau)
    &=\log\int
    p_\theta(\sD_\tau\mid\rvz_\tau)
    p(\rvz_\tau)\,\mathrm d\rvz_\tau
    \nonumber\\
    &=\log\int
    q_\phi(\rvz_\tau\mid\sD_\tau)
    \frac{
        p_\theta(\sD_\tau\mid\rvz_\tau)p(\rvz_\tau)
    }{
        q_\phi(\rvz_\tau\mid\sD_\tau)
    }
    \,\mathrm d\rvz_\tau
    \nonumber\\
    &\geq
    \mathbb E_{q_\phi(\rvz_\tau\mid\sD_\tau)}
    \!\left[
        \log p_\theta(\sD_\tau\mid\rvz_\tau)
        +\log p(\rvz_\tau)
        -\log q_\phi(\rvz_\tau\mid\sD_\tau)
    \right]
    \nonumber\\
    &=\mathbb E_{q_\phi(\rvz_\tau\mid\sD_\tau)}
    \!\left[\log p_\theta(\sD_\tau\mid\rvz_\tau)\right]
    -D_{\mathrm{KL}}\!\left(
        q_\phi(\rvz_\tau\mid\sD_\tau)
        \,\middle\|\,
        p(\rvz_\tau)
    \right)
    =:\mathcal E_\tau.
    \label{eq:joint_elbo_derivation}
\end{align}

\noindent\textbf{2. Gaussian-prior conditional ELBO.}
During training, we partition the realizations into disjoint sets
$\sD_\tau=\sC_\tau\cup\sT_\tau$, with
$|\sC_\tau|=K_C$, $|\sT_\tau|=K_T$, and $K_\tau=K_C+K_T$.
Under the same Gaussian-prior model, the context induces
$p_\theta(\rvz_\tau\mid\sC_\tau)$, and the conditional evidence is
\begin{equation}
    p_\theta(\sT_\tau\mid\sC_\tau)
    =
    \int
    p_\theta(\sT_\tau\mid\rvz_\tau)
    p_\theta(\rvz_\tau\mid\sC_\tau)
    \,\mathrm d\rvz_\tau .
    \label{eq:true_target_marginal}
\end{equation}
Using $q_\phi(\rvz_\tau\mid\sC_\tau\cup\sT_\tau)$ gives
\begin{align}
    \log p_\theta(\sT_\tau\mid\sC_\tau)
    &\geq
    \mathbb E_{q_\phi(\rvz_\tau\mid\sC_\tau\cup\sT_\tau)}
    \!\left[
        \log p_\theta(\sT_\tau\mid\rvz_\tau)
    \right]
    \nonumber\\
    &\quad -
    D_{\mathrm{KL}}
    \!\left(
        q_\phi(\rvz_\tau\mid\sC_\tau\cup\sT_\tau)
        \,\middle\|\,
        p_\theta(\rvz_\tau\mid\sC_\tau)
    \right)
    =:
    \mathcal E_\tau^{\mathrm{cond}} .
    \label{eq:exact_conditional_elbo}
\end{align}

\noindent\textbf{3. Learned conditional ELBO (our objective).}
Our model does not use the Gaussian-prior posterior
$p_\theta(\rvz_\tau\mid\sC_\tau)$ as its conditional prior. Instead, it uses
the learned encoder distribution $q_\phi(\rvz_\tau\mid\sC_\tau)$, defining
\begin{equation}
    \widetilde p_{\phi,\theta}(\sT_\tau\mid\sC_\tau)
    =\int
    p_\theta(\sT_\tau\mid\rvz_\tau)
    q_\phi(\rvz_\tau\mid\sC_\tau)
    \,\mathrm d\rvz_\tau.
\end{equation}
Applying Jensen's inequality with
$q_\phi(\rvz_\tau\mid\sC_\tau\cup\sT_\tau)$ gives the objective used in
\Eqref{eq:training_objective}:
\begin{align}
    \log\widetilde p_{\phi,\theta}(\sT_\tau\mid\sC_\tau)
    &\geq
    \mathbb E_{q_\phi(\rvz_\tau\mid\sC_\tau\cup\sT_\tau)}
    \!\left[\log p_\theta(\sT_\tau\mid\rvz_\tau)\right]
    \nonumber\\
    &\quad-D_{\mathrm{KL}}\!\left(
        q_\phi(\rvz_\tau\mid\sC_\tau\cup\sT_\tau)
        \,\middle\|\,
        q_\phi(\rvz_\tau\mid\sC_\tau)
    \right)
    =:\mathcal E_\tau^{\mathrm{amortized}}.
\end{align}
Thus, $\mathcal E_\tau^{\mathrm{amortized}}$ is a valid ELBO for the learned
conditional evidence. If
$q_\phi(\rvz_\tau\mid\sC_\tau)=p_\theta(\rvz_\tau\mid\sC_\tau)$, the second
and third ELBOs coincide.

\section{Synthetic Experimental Details}
\label{app:synthetic_details}

\subsection{Data Generation and Dataset Splits}
\label{app:experimental_configuration}

We generate 5,500 tasks on $[0,T]$ with $T=10$. Each task has one latent intensity and 16 iid realizations. As in Section~\ref{sec:synthetic_setup},
\begin{equation}
    f_\tau\sim\operatorname{GP}(0,C_{\mathrm{RBF}}),
    \qquad
    \lambda_\tau(t)=20\operatorname{softplus}(f_\tau(t)),
    \label{eq:appendix_synthetic_hierarchical}
\end{equation}
where $C_{\mathrm{RBF}}(t,t')=\sigma_f^2\exp(-(t-t')^2/(2\ell^2))$,
with signal variance $\sigma_f^2=1$ and lengthscale $\ell=2$. We sample each GP on
an equally spaced grid of $G=300$. The same grid is used for intensity evaluation.

The synthetic dataset contains 4,500 training, 500 validation, and 500 test
tasks. Splits are disjoint at the task level. Table~\ref{tab:synthetic_split_stats}
summarizes event counts using all 16 realizations of every task. Event time standardization is estimated from the
training split and held fixed for validation, testing, and OOD evaluation.

\begin{table}[htbp]
\centering
\caption{Synthetic dataset statistics.}
\label{tab:synthetic_split_stats}
\small
\begin{tabular}{@{}lrrrrr@{}}
\toprule
Split & Tasks & Sequences & Events & Events/sequence & Range \\
\midrule
Train & 4,500 & 72,000 & 11,625,135 & $161.46\pm67.80$ & 15--484 \\
Validation & 500 & 8,000 & 1,265,973 & $158.25\pm68.72$ & 7--441 \\
Test & 500 & 8,000 & 1,275,955 & $159.49\pm64.91$ & 13--444 \\
\bottomrule
\end{tabular}
\end{table}

\subsection{Training and Checkpoint Selection}
\label{app:synthetic_selection}

Each update samples a batch of 256 training tasks. We draw $K_C$ once
per batch from the mixture specified in Section~\ref{sec:synthetic_setup}
and set $K_T=K_{\mathrm{train}}-K_C$. Each task's realizations are
partitioned into $K_C$ context and $K_T$ target realizations.
We train each $K_{\mathrm{train}}\in\{4,8,16\}$ configuration for
20,000 updates with seeds 42--45, using AdamW with initial learning
rate $10^{-3}$, weight decay $10^{-4}$, and cosine learning rate decay.

We save a checkpoint every 1,000 updates. After training, all 20
checkpoints are evaluated on the 500 validation tasks. We hold out each
realization in turn and use the remaining realizations as context.
The checkpoint with the highest validation HLL is selected separately
for each training seed. Neither ground-truth intensity labels nor test
tasks are used for selection.

Table~\ref{tab:synthetic_checkpoint_selection} reports the validation HLL
of every saved checkpoint for seed 42. For all three values of
$K_{\mathrm{train}}$, the checkpoint at 20,000 updates achieves the highest
validation HLL and is therefore selected.
\begin{table}[htbp]
    \centering
    \caption{Validation HLL across saved checkpoints for
    seed 42. Values are shown to two decimal places.}
    \label{tab:synthetic_checkpoint_selection}
    \small
    \setlength{\tabcolsep}{12pt}
    \begin{tabular}{rrrr}
        \toprule
        & \multicolumn{3}{c}{Validation HLL $\uparrow$} \\
        \cmidrule(lr){2-4}
        Update & $K_{\mathrm{train}}=4$ & $K_{\mathrm{train}}=8$
        & $K_{\mathrm{train}}=16$ \\
        \midrule
        1,000  & 308.32 & 308.69 & 309.38 \\
        2,000  & 310.11 & 309.21 & 310.71 \\
        3,000  & 310.76 & 309.61 & 310.85 \\
        4,000  & 310.62 & 309.89 & 310.90 \\
        5,000  & 310.81 & 310.11 & 311.02 \\
        6,000  & 310.84 & 310.28 & 311.04 \\
        7,000  & 310.90 & 310.28 & 310.85 \\
        8,000  & 310.94 & 310.26 & 311.08 \\
        9,000  & 310.98 & 310.31 & 311.00 \\
        10,000 & 310.99 & 310.36 & 311.18 \\
        11,000 & 310.91 & 310.38 & 311.14 \\
        12,000 & 310.96 & 310.37 & 311.19 \\
        13,000 & 311.00 & 310.41 & 311.19 \\
        14,000 & 311.00 & 310.40 & 311.20 \\
        15,000 & 311.01 & 310.40 & 311.20 \\
        16,000 & 311.01 & 310.41 & 311.21 \\
        17,000 & 311.02 & 310.41 & 311.22 \\
        18,000 & 311.02 & 310.42 & 311.22 \\
        19,000 & 311.03 & 310.43 & 311.22 \\
        \textbf{20,000} & \textbf{311.03} & \textbf{310.43} & \textbf{311.23} \\
        \bottomrule
    \end{tabular}
\end{table}

To estimate HLL in \Eqref{eq:heldout_hll}, we draw
$\rvz_\tau^{(j)}\sim q_\phi(\rvz_\tau\mid\sD_{\tau,-h})$
for each held-out realization $h$. Its point process log likelihood
under sample $j$ is
\begin{equation}
    \ell_{\tau,h}^{(j)}
    =\sum_{t_i\in\sS_{\tau,h}}
    \log\lambda_\theta(t_i\mid\rvz_\tau^{(j)})
    -\int_0^T\lambda_\theta(t\mid\rvz_\tau^{(j)})\,\mathrm dt.
    \label{eq:heldout_sample_loglik}
\end{equation}
For $s\in\{\mathrm{val},\mathrm{test}\}$, the Monte Carlo estimate is
\begin{equation}
    \widehat{\operatorname{HLL}}_s
    =\frac{1}{|\mathcal D_s|}
    \sum_{\sD_\tau\in\mathcal D_s}\frac{1}{|\sD_\tau|}
    \sum_{h=1}^{|\sD_\tau|}
    \log\!\left[
        \frac{1}{n_{\mathrm{MC}}}
        \sum_{j=1}^{n_{\mathrm{MC}}}\exp\{\ell_{\tau,h}^{(j)}\}
    \right].
    \label{eq:heldout_hll_mc}
\end{equation}
We compute the inner log average using log-mean-exp with
$n_{\mathrm{MC}}=300$.

\subsection{Evaluation Protocol and Metrics}
\label{app:synthetic_metrics}

To evaluate generalization to numbers of realizations not seen during training, we retain the 16 realizations of each of the 500 test tasks and generate 48 additional realizations from the same underlying intensity.
Evaluation uses the first $K_{\mathrm{test}}\in\{1,2,4,8,16,32,64\}$
realizations, giving nested contexts as replication increases.
The underlying test intensities and contexts are shared across methods.

For $n_{\mathrm{test}}=500$ test tasks, let $\lambda_{\tau,g}$ be the
ground-truth intensity at grid point $t_g$, $g=1,\ldots,G$.
Let $\rvz_\tau^{(m)}$, $m=1,\ldots,n_{\mathrm{MC}}$, denote
$n_{\mathrm{MC}}=300$ draws from $q_\phi(\rvz_\tau\mid\sD_\tau)$, and set
$\lambda_{\tau,g}^{(m)}=\lambda_\theta(t_g\mid\rvz_\tau^{(m)})$.
The posterior mean intensity is
\[
\widehat\lambda_{\tau,g}
=\frac{1}{n_{\mathrm{MC}}}\sum_{m=1}^{n_{\mathrm{MC}}}\lambda_{\tau,g}^{(m)}.
\]
Let $\ell_{\tau,g}$ and $u_{\tau,g}$ be the pointwise 2.5\%
and 97.5\% sample quantiles. The task-averaged metrics are
\begin{align}
\operatorname{RMSE}
&=\frac{1}{n_{\mathrm{test}}}\sum_{\tau=1}^{n_{\mathrm{test}}}
  \sqrt{\frac{1}{G}\sum_{g=1}^{G}
  (\widehat\lambda_{\tau,g}-\lambda_{\tau,g})^2},
  \nonumber\\
\operatorname{C95}
&=\frac{1}{n_{\mathrm{test}}}\sum_{\tau=1}^{n_{\mathrm{test}}}\left[
  \frac{1}{G}\sum_{g=1}^{G}
  \mathbf 1\!\left\{
  \ell_{\tau,g}\leq\lambda_{\tau,g}\leq u_{\tau,g}
  \right\}\right],
  \nonumber\\
\operatorname{IW95}
&=\frac{1}{n_{\mathrm{test}}}\sum_{\tau=1}^{n_{\mathrm{test}}}\left[
  \frac{1}{G}\sum_{g=1}^{G}(u_{\tau,g}-\ell_{\tau,g})\right].
\label{eq:synthetic_metrics}
\end{align}

\subsection{Additional Training-Replication Results}
\label{app:additional_ktrain_results}

Table~\ref{tab:synthetic_cross_k_additional} reports the corresponding results
for models trained with $K_{\mathrm{train}}\in\{8,16\}$. Both configurations
show lower RMSE and narrower CIs as $K_{\mathrm{test}}$
increases, while empirical coverage remains broadly stable. Compared with the
$K_{\mathrm{train}}=4$ model in Table~\ref{tab:synthetic_cross_k}, these models
achieve lower RMSE when many test realizations are available, but slightly
higher RMSE when $K_{\mathrm{test}}$ is small.

\begin{table}[htbp]
    \centering
    \caption{Evidence accumulation for models trained with
    $K_{\mathrm{train}}\in\{8,16\}$. Results are averaged over four seeds.}
    \label{tab:synthetic_cross_k_additional}
    \small
    \begin{tabular}{ccccccc}
        \toprule
        & \multicolumn{2}{c}{\textbf{RMSE} $\downarrow$}
        & \multicolumn{2}{c}{\textbf{C95 (\%)}}
        & \multicolumn{2}{c}{\textbf{IW95}} \\
        \cmidrule(lr){2-3}\cmidrule(lr){4-5}\cmidrule(lr){6-7}
        $K_{\mathrm{test}}$ & $K_{\mathrm{train}}=8$ & $K_{\mathrm{train}}=16$
        & $K_{\mathrm{train}}=8$ & $K_{\mathrm{train}}=16$
        & $K_{\mathrm{train}}=8$ & $K_{\mathrm{train}}=16$ \\
        \midrule
        1  & $2.75_{\pm 0.01}$ & $2.81_{\pm 0.01}$ & $92.23_{\pm 0.16}$ & $89.81_{\pm 2.62}$ & $9.60_{\pm 0.03}$ & $9.14_{\pm 0.61}$ \\
        2  & $2.03_{\pm 0.01}$ & $2.06_{\pm 0.01}$ & $92.66_{\pm 0.10}$ & $90.64_{\pm 2.35}$ & $7.12_{\pm 0.01}$ & $6.77_{\pm 0.44}$ \\
        4  & $1.56_{\pm 0.01}$ & $1.56_{\pm 0.01}$ & $91.90_{\pm 0.16}$ & $89.73_{\pm 1.77}$ & $5.37_{\pm 0.01}$ & $5.04_{\pm 0.24}$ \\
        8  & $1.19_{\pm 0.01}$ & $1.18_{\pm 0.01}$ & $92.78_{\pm 0.26}$ & $90.94_{\pm 0.57}$ & $4.23_{\pm 0.01}$ & $3.89_{\pm 0.06}$ \\
        16 & $0.93_{\pm 0.01}$ & $0.91_{\pm 0.01}$ & $93.74_{\pm 0.31}$ & $92.40_{\pm 0.74}$ & $3.54_{\pm 0.01}$ & $3.18_{\pm 0.07}$ \\
        32 & $0.79_{\pm 0.01}$ & $0.75_{\pm 0.01}$ & $93.46_{\pm 0.20}$ & $92.95_{\pm 1.33}$ & $3.12_{\pm 0.01}$ & $2.76_{\pm 0.13}$ \\
        64 & $0.71_{\pm 0.02}$ & $0.66_{\pm 0.02}$ & $93.12_{\pm 0.22}$ & $93.19_{\pm 1.79}$ & $2.89_{\pm 0.01}$ & $2.53_{\pm 0.17}$ \\
        \bottomrule
    \end{tabular}
\end{table}

\subsection{Baseline Implementations}
\label{app:synthetic_baselines_runtime}

\paragraph{PSTH.}
We divide pooled bin counts by the number of input realizations and the
bin width. We select the bin width on the 500 validation tasks using
per-task intensity RMSE from four input realizations. The selected width
$1.0$ is fixed for test evaluation.

\paragraph{KDE.}
We use a boundary-normalized Gaussian kernel and select its bandwidth
using the same validation RMSE. The selected bandwidth $0.5$ is
fixed for four-realization test evaluation. Table~\ref{tab:synthetic_classical_selection}
gives both validation searches.

\begin{table}[t]
    \centering
    \caption{Validation selection for deterministic baselines on the synthetic dataset.}
    \label{tab:synthetic_classical_selection}
    \small
    \begin{tabular}{rrrr}
        \toprule
        PSTH bin width & Validation RMSE & KDE bandwidth & Validation RMSE \\
        \midrule
        0.05 & 8.67 & 0.05 & 4.65 \\
        0.10 & 6.13 & 0.10 & 3.32 \\
        0.20 & 4.33 & 0.20 & 2.42 \\
        0.30 & 4.93 & 0.30 & 2.08 \\
        0.50 & 2.80 & 0.50 & \textbf{1.87} \\
        0.75 & 5.78 & 0.75 & 2.03 \\
        1.00 & \textbf{2.40} & 1.00 & 2.40 \\
        \midrule
        \multicolumn{1}{l}{Selected bin width} & 1.00 & \multicolumn{1}{l}{Selected bandwidth} & 0.50 \\
        \multicolumn{1}{l}{Test RMSE} & 2.45 & \multicolumn{1}{l}{Test RMSE} & 1.91 \\
        \bottomrule
    \end{tabular}
\end{table}

\paragraph{LGCP.}
The LGCP baseline is fitted directly to each test task using an RBF prior, an exponential link, 10 inducing
points, and 10,000 variational optimization steps per test task. We use a Gaussian variational distribution over the inducing variables and jointly optimize the GP mean, lengthscale, and signal variance. Metrics use 300 intensity samples.

\paragraph{FIM-PP.}
We use the released
\href{https://huggingface.co/FIM4Science/fim-pp}{\texttt{FIM4Science/fim-pp}}
checkpoint without fine-tuning. Its context contains the first four realizations, and its inference sequence is the fifth realization. It therefore
receives an additional observed sequence compared with the other baselines. 

Table~\ref{tab:synthetic_compute} summarizes the compute environment and
training or fitting budget for Ours, LGCP, and FIM-PP.

\begin{table}[htbp]
    \centering
    \caption{Synthetic compute and implementation record.}
    \label{tab:synthetic_compute}
    \small
    \begin{tabular*}{\textwidth}{@{\extracolsep{\fill}}llll@{}}
        \toprule
        Method & Hardware & Training time & Checkpoint / fit budget \\
        \midrule
        Ours, $K_{\mathrm{train}}=4$ & NVIDIA GH200 & 2,969 s & 20,000 updates \\
        Ours, $K_{\mathrm{train}}=8$ & NVIDIA GH200 & 3,497 s & 20,000 updates \\
        Ours, $K_{\mathrm{train}}=16$ & NVIDIA GH200 & 4,305 s & 20,000 updates \\
        LGCP & NVIDIA GH200 & --- & 10 inducing points; 10,000 steps per task \\
        FIM-PP & NVIDIA GH200 & Pretrained & Released checkpoint \\
        \bottomrule
    \end{tabular*}
\end{table}

\subsection{Out-of-Distribution Test}
\label{app:synthetic_ood}

Both OOD evaluations use the $K_{\mathrm{train}}=8$ models and $K_{\mathrm{test}}=8$.
Each test set contains 500 tasks.

\paragraph{OOD1: constant intensity.}
We draw a constant intensity uniformly
from $[5,20]$ and generate 8 realizations.

\paragraph{OOD2: GP--constant--linear intensity.}
We draw transition times $b_1\sim\operatorname{Unif}(2,4)$
and $b_2\sim\operatorname{Unif}(6,8)$ and set $c$ to the original GP
intensity at $b_1$. The transformed intensity is
\begin{equation}
\lambda_\tau^{\mathrm{OOD2}}(t)=
\begin{cases}
\lambda_\tau(t), & 0\leq t<b_1,\\
c, & b_1\leq t<b_2,\\
c[1+r(t-b_2)], & b_2\leq t\leq T,
\end{cases}
\label{eq:synthetic_ood2}
\end{equation}
where $r\sim\operatorname{Unif}(-0.9/(T-b_2),\,1.5/(T-b_2))$.
Thus the final intensity is between $0.1c$ and $2.5c$.

\begin{figure}[htbp]
    \centering
    \includegraphics[width=0.45\textwidth]{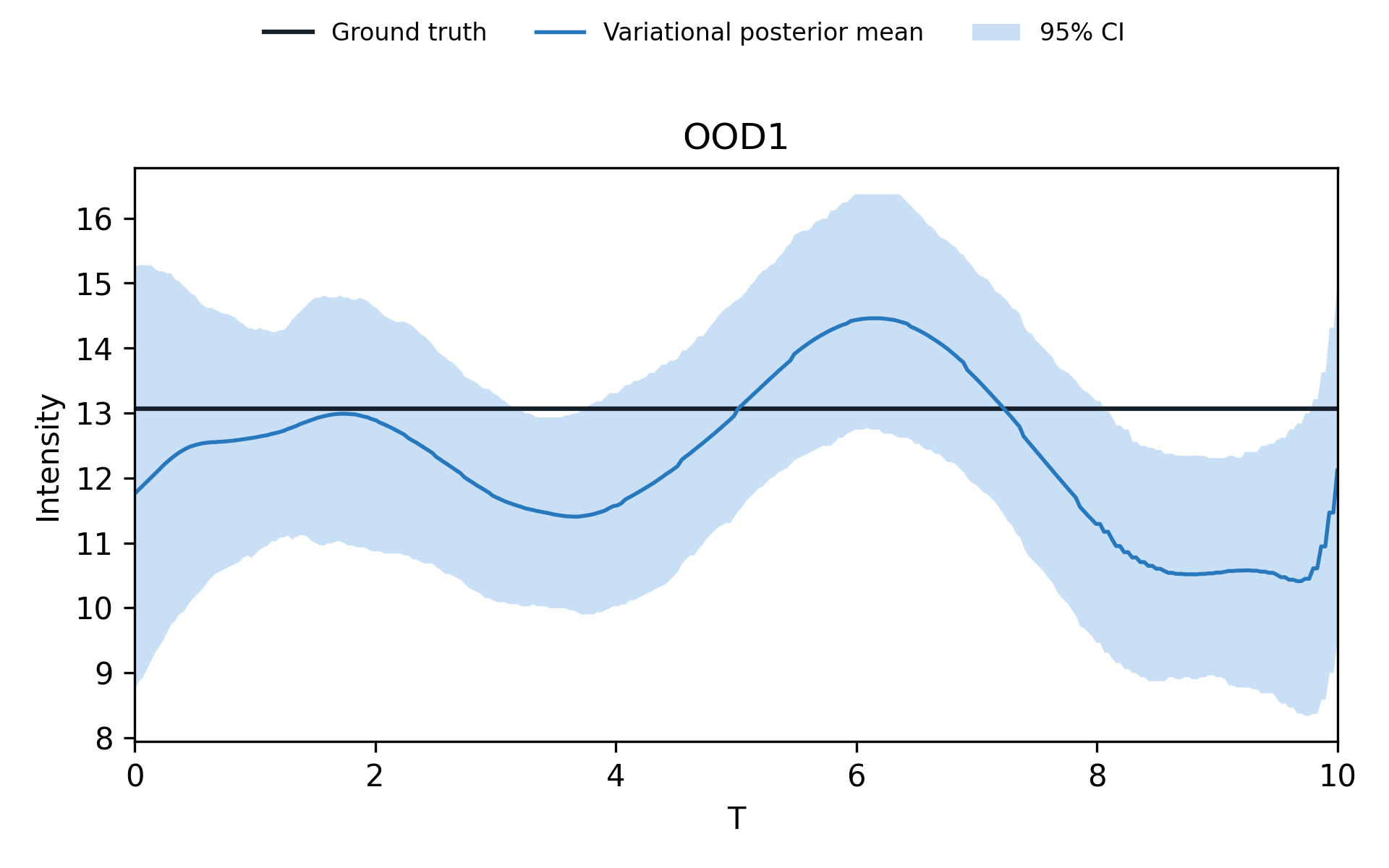}\qquad
    \includegraphics[width=0.45\textwidth]{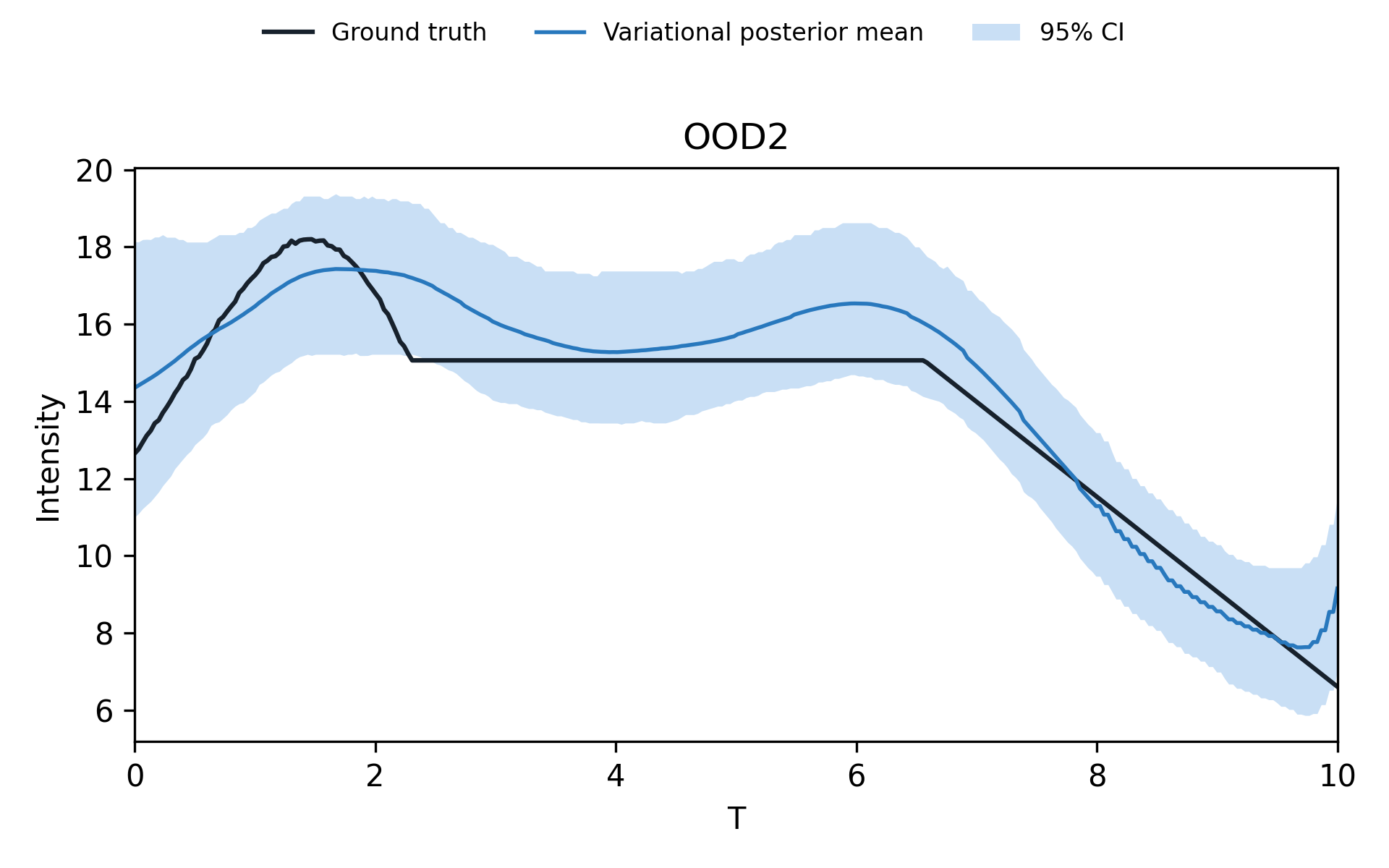}
    \caption{OOD intensity recovery at $K_{\mathrm{test}}=8$ for OOD1 (left)
    and OOD2 (right).}
    \label{fig:ood_examples}
\end{figure}

Across four training seeds, OOD1 attains C95 of $95.72$, while OOD2 has
lower coverage of $91.99$ and higher RMSE (Table~\ref{tab:ood_results}).
Representative tasks are shown in Figure~\ref{fig:ood_examples}.

\begin{table}[htbp]
    \centering
    \caption{OOD results. Results are averaged over four training seeds.}
    \label{tab:ood_results}
    \small
    \begin{tabular}{lccc}
        \toprule
        Test set & RMSE  & C95 (\%) & IW95 \\
        \midrule
        OOD1 & $0.89_{\pm 0.01}$ & $95.72_{\pm 0.10}$ & $3.77_{\pm 0.01}$ \\
        OOD2 & $1.23_{\pm 0.02}$ & $91.99_{\pm 0.17}$ & $4.25_{\pm 0.01}$ \\
        \bottomrule
    \end{tabular}
\end{table}

\section{Real-Data Protocol and Metrics}
\label{app:real_data_details}

\subsection{Data Construction}
\label{app:real_data_construction}

We use the Divvy bicycle-share trip records from the City of Chicago Open
Data Portal for 2013--2019.
Records include origin and
destination stations and trip timestamps; we use stations
and departure times. It contains 21,242,740 trips. We treat each
trip start as an event and group starts by origin station and
day, producing 951,508 station-day sequences.

A task is a station--year--month--weekday group with at least four
observed days. We retain the first four days chronologically as its $K=4$
realizations, yielding 194,334 candidate tasks. We first compare the four total
event counts under an equal rate Poisson null using a Pearson chi-square test. We retain 148,151 tasks.

We divide each day into 10 equal-width time bins and use a Pearson test to assess whether the four days share the same normalized temporal profile. Tasks with \(p\geq0.01\) are retained.

We select the top 10\% by total events across the four realizations. This yields
13,468 tasks and a cutoff of 163 events per task. Stations are then assigned
disjointly to training, validation, and test splits, giving 10,774/1,347/1,347 tasks. Finally, we sample half of the tasks
within each split and obtain the final dataset. Table~\ref{tab:divvy_data_flow}
summarizes the resulting dataset.

\begin{table}[H]
    \centering
    \caption{Divvy dataset after preprocessing.}
    \label{tab:divvy_data_flow}
    \small
    \setlength{\tabcolsep}{5pt}
    \begin{tabular}{lrrrrr}
        \toprule
        Split & Tasks & Sequences & Stations & Events & Events/task \\
        \midrule
        Train      & 5,387 & 21,548 & 182 & 1,437,806 & 163--1,534 \\
        Validation & 674   & 2,696  & 21  & 170,078   & 163--1,018 \\
        Test       & 674   & 2,696  & 21  & 184,032   & 163--1,154 \\
        \bottomrule
    \end{tabular}
\end{table}

\subsection{Training and Checkpoint Selection}
\label{app:real_data_training}

Our model is trained on the 5,387 tasks in
$\mathcal D_{\mathrm{train}}$. Feature normalization statistics
are computed from the training split only. We use AdamW with initial learning
rate $10^{-3}$, cosine decay, weight decay $10^{-4}$, and batches of 256 tasks for 20,000 updates. 
For each task, the context size $K_C$ is sampled from
$\tfrac12\operatorname{Uniform}\{2,3\}+\tfrac12\operatorname{Uniform}\{1,2,3\}$.
We save a checkpoint every
1,000 updates. After training, all 20 checkpoints are evaluated on the 674 tasks
in $\mathcal D_{\mathrm{val}}$ using the four-fold HLL evaluation in
Appendix~\ref{app:synthetic_selection}, with $n_{\mathrm{MC}}=300$ latent samples. The checkpoint with the
highest validation HLL is selected for each seed.
Table~\ref{tab:divvy_validation_hll} reports the validation HLL
of all saved checkpoints for seeds 42--45.

\begin{table}[htbp]
    \centering
    \caption{Validation HLL across saved checkpoints on Divvy.
    Bold indicates the best checkpoint for each seed, determined before rounding.}
    \label{tab:divvy_validation_hll}
    \small
    \setlength{\tabcolsep}{10pt}
    \begin{tabular}{rrrrr}
        \toprule
        & \multicolumn{4}{c}{Validation HLL $\uparrow$} \\
        \cmidrule(lr){2-5}
        Update & Seed 42 & Seed 43 & Seed 44 & Seed 45 \\
        \midrule
        1,000 & 82.87 & 82.63 & 81.67 & 81.95 \\
        2,000 & 83.62 & 83.61 & 83.30 & 82.95 \\
        3,000 & 83.55 & 83.75 & 83.72 & 83.59 \\
        4,000 & 83.45 & 83.71 & 83.83 & 83.47 \\
        5,000 & 83.99 & 84.00 & 83.87 & 83.99 \\
        6,000 & 83.84 & 83.87 & 84.03 & 83.97 \\
        7,000 & 84.14 & 84.10 & 83.97 & 84.23 \\
        8,000 & 84.19 & 84.18 & 83.97 & 84.16 \\
        9,000 & 84.25 & 84.16 & 84.19 & 83.77 \\
        10,000 & 84.21 & 84.27 & 84.26 & 84.22 \\
        11,000 & 84.23 & 84.23 & 84.19 & 84.23 \\
        12,000 & 84.14 & 84.25 & 84.22 & 84.25 \\
        13,000 & 84.28 & 84.30 & 84.25 & 84.26 \\
        14,000 & 84.27 & 84.32 & 84.31 & 84.27 \\
        15,000 & 84.30 & 84.29 & 84.30 & 84.29 \\
        16,000 & 84.29 & 84.31 & 84.30 & 84.30 \\
        17,000 & 84.31 & \textbf{84.34} & 84.31 & 84.31 \\
        18,000 & \textbf{84.32} & 84.34 & \textbf{84.33} & \textbf{84.32} \\
        19,000 & 84.32 & 84.33 & 84.32 & 84.31 \\
        20,000 & 84.32 & 84.34 & 84.33 & 84.31 \\
        \bottomrule
    \end{tabular}
\end{table}

\subsection{Baseline Selection and Implementation}
\label{app:real_data_baselines}

We select the PSTH bin width and KDE bandwidth by four-fold
leave-one-realization-out likelihood on all 674 validation tasks. Both searches
use the candidate grid $\{0.05,0.1,0.2,0.3,0.5,0.75,1.0\}$. The selected values
are 1.0 for PSTH and 0.3 for KDE. For both methods, HLL is computed using their point estimates of the intensity.

LGCP is fit independently in each fold using 10 inducing points and
10,000 optimization steps. Its HLL is computed by averaging
the point process likelihood over 300 samples from its variational intensity
distribution and then taking the logarithm, following the same Monte Carlo
evaluation as our method.

FIM-PP requires one inference sequence in addition to its context. For each
held-out realization, we therefore run FIM-PP three times. Each visible
realization serves once as the inference sequence, while the other two form the
context. We score the held-out realization under each resulting intensity,
average the three likelihoods, and then take the logarithm.

\end{document}